\documentclass[lettersize,journal]{IEEEtran}
\usepackage{amsmath,amsfonts}
\usepackage{algorithmic}
\usepackage{algorithm}
\usepackage{array}
\usepackage[caption=false,font=normalsize,labelfont=sf,textfont=sf]{subfig}
\usepackage{textcomp}
\usepackage{stfloats}
\usepackage{url}
\usepackage{verbatim}
\usepackage{graphicx}
\usepackage{float}
\usepackage{cite}

\usepackage{booktabs}
\usepackage{arydshln}
\usepackage{float} 
\usepackage{multirow,multicol}
\usepackage{CJKutf8}
\newcommand{\STAB}[1]{\begin{tabular}{@{}c@{}}#1\end{tabular}}

\usepackage{tikz,tikz-qtree}
\usepackage{pgfplots}
\pgfplotsset{compat=1.14}
\definecolor{g-red}{HTML}{DB4437}
\definecolor{g-blue}{HTML}{4285F4}
\definecolor{g-green}{HTML}{0F9D58}
\definecolor{g-yellow}{HTML}{F4B400}
\definecolor{g-orange}{HTML}{FF9800}
\definecolor{g-grey}{HTML}{9E9E9E}

\definecolor{f1-y}{RGB}{255,240,193}
\definecolor{f1-o}{RGB}{245,191,157}
\definecolor{f1-b}{RGB}{176,205,234}

\usepackage{colortbl}

\begin{document}

\title{Bridging Cross-Lingual Gaps During Leveraging the Multilingual Sequence-to-Sequence Pretraining for Text Generation and Understanding}

\author{
        Changtong Zan$^\dagger$,
        Liang Ding$^\ddagger$,
        Li Shen, 
        Yu Cao,  \\
        Weifeng Liu$^\ddagger$,~\IEEEmembership{Senior Member,~IEEE}
        and~Dacheng Tao,~\IEEEmembership{Fellow,~IEEE}
\IEEEcompsocitemizethanks{
\IEEEcompsocthanksitem C. Zan and W. Liu are with College of Control Science and Engineering, China University of Petroleum (East China), Qingdao , China (e-mail: b20050011@s.upc.edu.cn; liuwf@upc.edu.cn).
\IEEEcompsocthanksitem L. Ding, L. Shen and D. Tao are with the JD Explore Academy at JD.com, Beijing, China (e-mail: dingliang1@jd.com; shenli100@jd.com, dacheng.tao@jd.com).

\IEEEcompsocthanksitem $^\dagger$Work was done when Changtong Zan interning at JD Explore Academy; $^\ddagger$Corresponding Authors: Liang Ding (e-mail: dingliang1@jd.com) and Weifeng Liu (e-mail: liuwf@upc.edu.cn ).
}
}

\maketitle

\begin{abstract}
For multilingual sequence-to-sequence pretrained language models (multilingual Seq2Seq PLMs), e.g. mBART~\cite{tacl_mBART}, the self-supervised pretraining task is trained on a wide range of monolingual languages, e.g. 25 languages from CommonCrawl\footnote{\url{https://commoncrawl.org/the-data/}}, while the downstream cross-lingual tasks generally progress on a bilingual language subset, e.g. English-German, making there exists the data discrepancy, namely \textit{domain discrepancy}, and cross-lingual learning objective discrepancy, namely \textit{task discrepancy}, between the pretraining and finetuning stages. To bridge the above cross-lingual domain and task gaps, we extend the vanilla pretrain-finetune pipeline with an extra code-switching restore task. Specifically, the first stage employs the self-supervised code-switching restore task as a pretext task, allowing the multilingual Seq2Seq PLMs to acquire some in-domain alignment information. And for the second stage, we fine-tune the model on downstream data normally. Experiments on both NLG evaluation (12 bilingual translation tasks, 30 zero-shot translation tasks, and 2 cross-lingual summarization tasks) and NLU evaluation (7 cross-lingual natural language inference tasks) show our model outperforms the strong baseline mBART with standard finetuning strategy, consistently. Analyses indicate our approach could narrow the Euclidean distance of cross-lingual sentence representations, and improve the model generalization with trivial computational cost. We release the code at:~\url{https://github.com/zanchangtong/CSR4mBART}.

\end{abstract}


\section{Introduction}
\IEEEPARstart{P}{retraining} has achieved tremendous success in natural language processing fields~\cite{9767637,9478264,9599397,9768828,Devlin2019BERTPO,Liu2019RoBERTaAR,Lample2019CrosslingualLM,Brown2020gpt} since they transfer the knowledge from large-scale of unlabeled data implicitly to the parameters of the pretrained models.
Inspired by their success, recent works~\cite{DBLP:conf/icml/SongTQLL19,tacl_mBART} attempt to design a unified multilingual sequence-to-sequence pretrained language model (multilingual Seq2Seq PLM) for cross-lingual downstream tasks, e.g. machine translation~\cite{transformer}, cross-lingual summarization~\cite{zhu-etal-2019-ncls}, and cross-lingual language understanding~\cite{xue-etal-2021-mt5}. 
Despite successfully moving the first step, the multilingual Seq2Seq PLMs encounter many challenges on cross-lingual NLU and NLG tasks~\cite{xue-etal-2021-mt5}. 
For example, with covering more language pairs, i.e. from 25 to 50, mBART50~\cite{tang2020multilingual} does not show any promising improvements (or even worse on 8 out of 24 languages) in given bilingual downstream tasks as reported by~\cite{tang2020multilingual}.

\begin{table}[t]
\setlength{\belowcaptionskip}{-0.2cm}
    \centering
    \caption{\textbf{Comparison of learning objectives between {multilingual Seq2Seq PLMs} \textit{pretraining} and \textit{finetuning}.} Multilingual Seq2Seq PLMs is trained on monolingual corpora of several languages, e.g. English ``[EN]'' and Chinese ``[ZH]'', and learns to predict the perturbed tokens, while finetuning on ZH$\rightarrow$EN parallel corpus is for cross-lingual translation. \underline{Underline} denotes perturbed text. Express the Chinese example sentence in English is ``The spokesperson holds a press conference once a week''.
    }
    \scalebox{1.0}{
    \begin{tabular}{p{1.2cm}p{6.3cm}}
    \toprule
    \multicolumn{2}{l}{\em Multilingual Seq2Seq PLM: $\mathcal-\log P(\bf{x}|\widetilde{\bf{x}})$} \\ 
    \hdashline
        \bf Source & Military \underline{{ }{ }{ }} Field Marshal Hussein \underline{{ }{ }{ }} in attendance. [EN] \\
        \bf Target & Military Field Marshal Hussein Tantawi was in attendance [EN] \\
        \bf Source &  \begin{CJK}{UTF8}{gbsn} 发言人 \underline{{ }{ }{ }} 一次新闻发布会\end{CJK} [ZH]\\
        \bf Target &  \begin{CJK}{UTF8}{gbsn} 发言人每周举行一次新闻发布会\end{CJK} [ZH] \\ 
          & ... ... ... [...]\\
                \hline
\multicolumn{2}{l}{\em finetuning on translation task: $\mathcal-\log P(\bf{y}|\bf{x})$} \\      
    \hdashline     
        \bf Source & \begin{CJK}{UTF8}{gbsn} 布什与沙龙举行了会谈\end{CJK} [ZH]\\
        \bf Target & Bush held a talk with Sharon. [EN]\\
    \bottomrule
    \end{tabular}}
    \label{tab:intro_example}
\end{table}

This deficiency may draw from the huge \textbf{\textsc{cross-lingual gap}} that exists between pretraining (PT) and finetuning (FT). Specifically, {we can categorize the cross-lingual gap into two aspects, which are as described by~\cite{wang2022understanding}}. (i) \textbf{``Domain discrepancy''}: 
PT are generally trained on monolingual data of multiple languages from a more general domain, e.g. 25 languages for mBART~\cite{tacl_mBART}, while the FT is mainly based on a bilingual subset, e.g. news domain English$\rightarrow$German; 
(ii) \textbf{``Task discrepancy''}: the task of PT is to utilize the monolingual data under a self-supervised pattern, such as denoising task for mBART, while the FT aims to generate the target-side language conditioned on the source-side language in a supervised fashion. 
Table~\ref{tab:intro_example} illustrates the detailed learning objective difference, multilingual Seq2Seq PLMs is pretrained with denoising tasks on monolingual corpora, e.g. English ``[EN]'', Chinese ``[ZH]'', while FT on the cross-lingual task, e.g. translation, highly relying on cross-lingual alignment information to accomplish the generation. 

{ Recent works that have been focused on designing advanced finetuning strategies for better utilizing the pretrained models can be divided into three categories, i.e. \textit{vocabulary}-level, \textit{task}-level, and \textit{model}-level. \cite{liu-etal-2021-bridging,hsu-etal-2020-efficient} show that bridging the vocabulary-level gap will narrow the subword and domain discrepancies between upstream and downstream tasks, thus improving the finetuning performance. \cite{aghajanyan-etal-2021-muppet,liu-etal-2019-multi} 
empirically validate the effectiveness of adopting task-level transferring strategies from the upstream to the downstream tasks via multi-task learning. 
Besides, \cite{Lee2020Mixout,li-etal-2021-multilingual,luo-etal-2021-veco} focus on the model-level adjusting, enhancing finetuning stability or training with less cost. 
Nevertheless, how to implement an effective finetuning strategy to bridge the cross-lingual gap and narrow it for Seq2Seq PLMs, including both domain discrepancy and task discrepancy, still remains an intractable problem with less attention.}

In this paper, we propose a two-stage finetuning strategy for multilingual Seq2Seq PLMs. Equipped with a code-switching restore task, it can effectively tune model on downstream cross-lingual generation and understanding tasks.
{ Our strategy sufficiently transfers the monolingual knowledge for multilingual Seq2Seq PLMs adapting to downstream cross-lingual tasks, thus improving the sub-optimal performance of finetuning caused by cross-lingual gap, i.e., domain discrepancy and task discrepancy.}
The first stage (\S\ref{subsec:csrestore}) is designed to inform the multilingual Seq2Seq PLMs about the downstream cross-lingual knowledge, by training it to denoise the code-switched samples for \textbf{mitigating the task discrepancy}. And we optimize these two tasks simultaneously to ensure the great potential to \textbf{mitigate the domain discrepancy}.
{ More specifically, we first derive the unsupervised translation vocabulary (in Table~\ref{fig: main CS-Annealing}) with downstream cross-lingual sentences following~\cite{artetxe2018acl}. 
Then, we introduce it by code-switching restore task. We perturb the downstream sentences by randomly replacing a certain percentage words with the semantic similar word of downstream another language.
For parameter estimation, we minimize the cross-entropy between the reconstructed output and the raw sample. To sufficiently transfer the knowledge from a multilingual Seq2Seq PLMs to the downstream cross-lingual tasks, we perform code-switching restore tasks in both the source- and target-side languages, and optimize these two tasks alternatively.
In the next stage, the standard finetuning is applied in the second step to derive the final model.
}

Experimental results on 12 \textit{bilingual translation} tasks, 30 \textit{zero-shot translation} tasks, 2 \textit{cross-lingual summarization} tasks, and 7 \textit{cross-lingual natural language inference} tasks show that our method could consistently outperform the standard finetuning.
{ We also conduct further analyses to provide some insights into our method and investigate how our approach bridging the cross-lingual gap:}
1) Our approach could narrow the Euclidean distance of cross-lingual sentence representations, especially for distant languages (e.g. 11.37 vs. 5.17 for Lt$\rightarrow$En), confirming our claim;
2) Our approach could improve the model generalization with a better low-frequency word translation accuracy;
and 3) Our approach only requires trivial computational costs, having great potential to be a universal plug-in strategy for any multilingual Seq2Seq PLMs.

{The main contributions of our work are three-fold:} 

\begin{itemize}
\item We propose a two-stage finetuning approach with the code-switching restore (CSR) task for multilingual Seq2Seq PLMs on cross-lingual generation and understanding tasks. 
\item Extensive experiments on 51 cross-language scenarios demonstrate that our proposed approach can effectively improve the performance of multilingual Seq2Seq PLMs. 
\item Additional analyses show that our approach indeed narrows the cross-lingual representation distance, benefiting the model generalization with trivial computational cost.
\end{itemize}

{ The subsequent paper is designed as follows. We discuss related works in Section~\ref{sec: related-works}. How we construct the code-switching restore task based two-stage finetuning is introduced in Section~\ref{sec: methods}. Experimental results and corresponding analyses are shown in Section~\ref{sec: exp} and Section~\ref{sec: discussion} respectively. Conclusions are described in Section~\ref{sec:conclusion}.}

\section{Related Work}
\label{sec: related-works}
{ Our work is inspired by three lines of research: \romannumeral1) sequence-to-sequence PLMs, \romannumeral2) code-switching based algorithms for pretrain-finetune paradigm and \romannumeral3) improving finetuning.}
\paragraph{Sequence-to-Sequence PLMs}
{ Seq2Seq PLMs bring remarkable improvements for natural language processing tasks, where large-scale unsupervised monolingual data is utilized via sequence-to-sequence self-supervised tasks. }
MASS~\cite{DBLP:conf/icml/SongTQLL19} takes a sentence with contiguous masked tokens as input and maps it to a sequence consisting of missing tokens. 
{ ``Text-to-Text Transfer Transformer'' (T5)~\cite{JMLR:v21:20-074} proposes unified text-to-text format to support a wide variety of English-language NLP tasks. And, mT5~\cite{xue-etal-2021-mt5} further extends T5 to 101 languages, verifying the universality of unified text-to-text format.}
BART~\cite{lewis-etal-2020-bart} analyzes the effect of different noise functions, shows remarkable performance on downstream monolingual understanding and generation tasks~\cite{zhong2022e2s2}. mBART~\cite{tacl_mBART} is a multilingual extension of BART that learns to reconstruct sentences whose orders are permutated and tokens are masked. While mBART and its variants~\cite{liu-etal-2021-copying,Liu2021OnTC,Zan2022OnTC} mainly focus on downstream bilingual translation tasks, mBART50~\cite{tang-etal-2021-multilingual} further extends mBART to 50 languages and analyzes the affect on multilingual translation scenarios.

However, the above pretraining tasks cannot offer models with cross-lingual alignment information. Therefore, bridging the cross-lingual gap for multilingual Seq2Seq PLMs between upstream and downstream tasks is worth investigating.

{\paragraph{Code-Switching Based Algorithms for Pretrain-Finetune Paradigm}
The basic idea of our work to address the cross-lingual gap between Seq2Seq multilingual PLMs and the finetuning process is the code-switching restore task, in which we train multilingual PLMs to predict the original sentence according to the code-switched one during finetuning. Several works also proposed to investigate benefits of code-switching in pretrain-finetune paradigm. CSP~\cite{yang-etal-2020-csp} focuses on the pretraining stage and uses unsupervised lexicon induction to construct a code-switching pretraining task, in which monolingual data is used, and a model is pretrained from scratch. 
{ CeMAT~\cite{li-etal-2022-universal} proposes aligned code-switching \& masking and dynamic dual-masking for pretraining the sequence-to-sequence transformer, achieving significant gains on both autoregressive translation and non-autoregressive translation tasks.
}
DICT-MLM~\cite{chaudhary2020dict} uses ground truth bilingual dictionaries in the code-switching task, being evaluated on cross-lingual transfer tasks rather than machine translation. 
{ Code-switching is also used in pretrained models of multilingual-multimodal scenarios. M3P~\cite{Ni_2021_CVPR} introduces the code-switching to effectively learn a universal representation and benefits the non-English tasks performance.}

Differing from above methods, our approach employs code-switched samples from downstream datasets for the denoising task as the first-stage finetuning, providing the standard finetuning with a better in-domain initialization with narrowed domain and task gaps.
}

\paragraph{Improving  Finetuning} 
Recent works have shown gains compared to directly optimize the downstream objective, based on modifications of finetuning in \textit{task}, \textit{model}, and \textit{vocabulary} levels. 

{ For task-level, new tasks are involved during finetuning~\cite{JMLR:v21:20-074, daume-iii-2007-frustratingly,khashabi-etal-2020-unifiedqa}. 
MTDNN~\cite{liu-etal-2019-multi} proves the efficiency of multi-task learning on top of the pretrained language models and evaluates on several NLU benchmarks. 
Nevertheless, no cross-lingual scenario is considered. Muppet~\cite{aghajanyan-etal-2021-muppet} scales up the type and number of intermediate tasks, which further verifies that multi-task learning could consistently bring benefits to various NLU tasks. 
Besides, several works also concentrate on further add additional regularization terms.
Inspired by trust-region theory, R3F~\cite{aghajanyan2021better} proposes to map samples in similar embeddings with small perturbations. Recadam~\cite{chen-etal-2020-recall} aims to recall the pretraining task by adding the difference of weights between the finetuned model and the original pretrained model to the loss function. }

{ Model-level works mainly focus on the issue of over-parameterization, where only part of pretrained parameters are initialized, or some components beneficial for downstream tasks are added~\cite{bowman-etal-2015-large,pmlr-v97-houlsby19a, liu2021unified}
Mixout~\cite{Lee2020Mixout} improves the model performance on the GLUE benchmark by randomly initializing part of parameters with pretrained weights. In addition, CHILD-TUNING~\cite{xu2021raise} gets a better ability of natural language understanding by masking the gradient of non-child networks during the backward process. LNA~\cite{li-etal-2021-multilingual} can also get the SOTA performance on two large-scale multilingual speech translation benchmarks, where only normalization layers and attention modules are trained. 
Besides finetuning PLMs with part of parameters, there also exits works concentrate on adjusting the Pre-training task, thus improving performance on specific target task type. 
VECO~\cite{luo-etal-2021-veco} adds a cross-attention module between two different languages for better capturing the cross-lingual information and performs better on downstream cross-lingual understanding tasks and translation tasks. 
}

{ For vocabulary-level, the vocabulary will be adjusted for more efficient inference to avoid possible mismatching for effective inference. \cite{liu-etal-2021-bridging} introduce an embedding generator to adapt the vocabulary, leading to a better performance on NLG tasks. \cite{hsu-etal-2020-efficient} achieve model acceleration without sacrifice for the accuracy based on model adjustment along with dictionary clipping.}

Our work belongs to the task-level approach, by introducing a pretext task with code-switching for finetuning multilingual Seq2Seq PLMs on cross-lingual generation and understanding. 
The difference between this work and previous ones is that we employ a code-switching restore task that samples in both languages from the following task is involved, ensuring the model is more adaptive to the downstream cross-lingual settings.

\section{Bridging Cross-Lingual Gaps} 
\label{sec: methods}
{ In this section, we present a two-stage finetuning approach with a code-switching restore task, which leads to better performance over standard finetuning both on cross-lingual generation tasks and cross-lingual understanding tasks.}

\begin{figure*}[t] 
    \centering
    \includegraphics[width=1\textwidth]{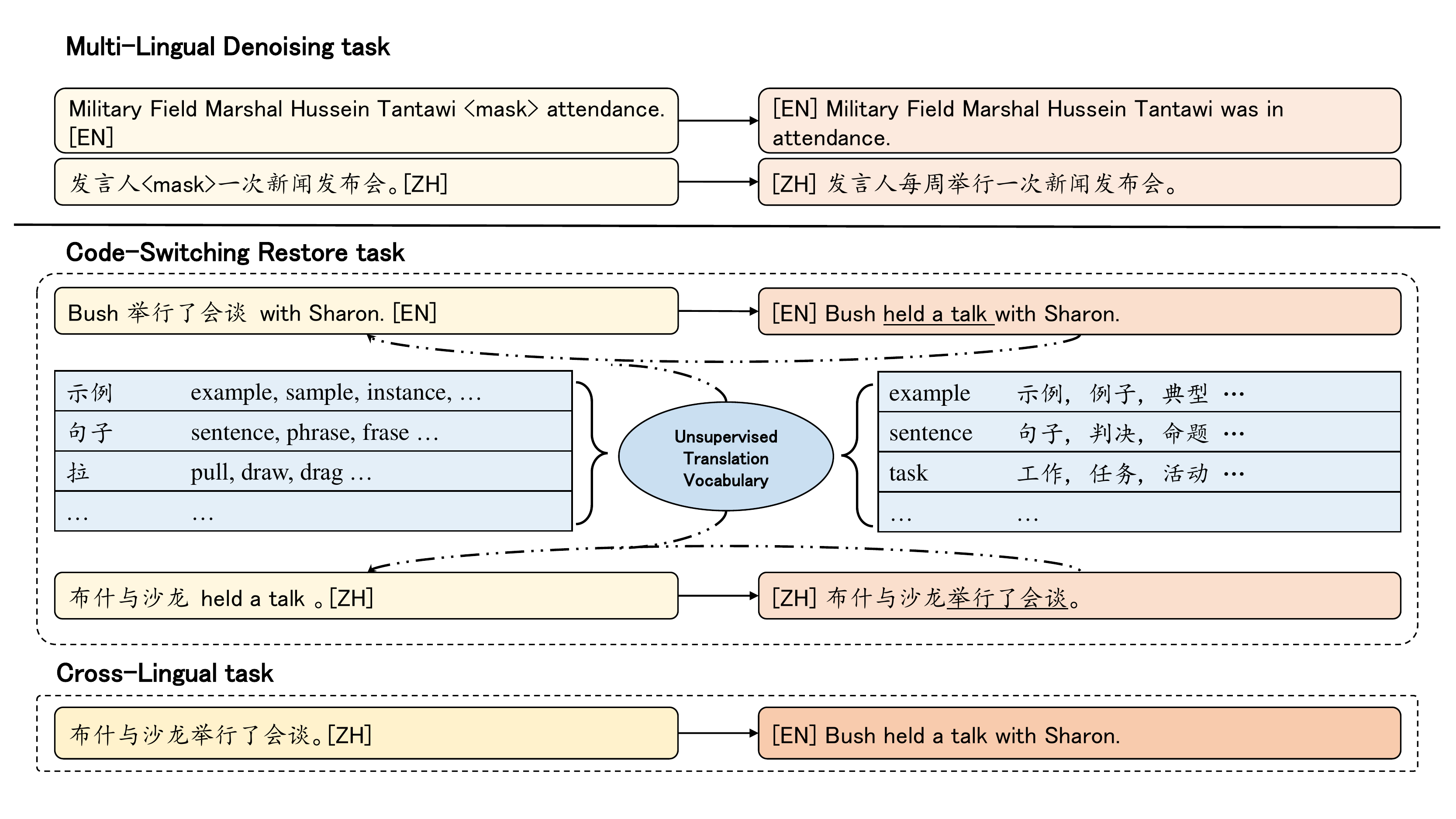}
    \caption{\textbf{The schematic comparison between \textit{pretraining stage} (top) of Multilingual Seq2Seq PLMs, \textit{standard finetuning stage} (bottom), and our proposed \textit{code-switching restore task} (middle).} 
    The rectangles colored with \textbf{\textcolor{f1-y}{yellow}}, and \textbf{\textcolor{f1-o}{orange}} represent the source and target language, respectively. 
    And areas colored with \textbf{\textcolor{f1-b}{blue}} mean our derived cross-lingual lexical alignments with downstream data.}
    \label{fig: main CS-Annealing}
\end{figure*} 

\subsection{Preliminary}
{ 
\paragraph{Sequence-to-Sequence Model based Cross-lingual Text Generation}
Sequence-to-sequence model, transformer, has achieved remarkable performance on various cross-lingual text generation tasks, which models source language sentences with bidirectional attention based encoder and generates target language representations with decoder based the encoder output. For a sample $S_i = \left ( X_i, Y_i \right ) = \left ({x}_1,..,{x}_n; {y}_1,..,{y}_m\right )$,  
the training of cross-lingual generation usually follows a teacher force fashion, maximizing the conditional generation objective over the training data $\left ( X, Y \right )$:
\begin{eqnarray}
\begin{split}
\label{eq:x-generation}
\mathcal{L}_\mathrm{gen}
&=\sum_{{X_i, Y_i}\in (X, Y)}-\log P(Y_i|X_i) \\
&=\sum_{{X_i, Y_i}\in (X Y)}\sum^{m}_{{j=1}}-\log P(y_j|y_1,...,y_{j-1}, X_i)
\end{split}
\end{eqnarray}
}

\paragraph{Denoising Task for Multilingual Sequence-to-Sequence Pretraining}
The denoising pretrain task learns a function that directly maps the sentence with noises to the original sentence, as shown at the top of Figure~\ref{fig: main CS-Annealing}. We adopt  mBART~\cite{tacl_mBART} as the backbone, which uses a standard sequence-to-sequence Transformer~\cite{transformer}, pretrained on the CC-25\footnote{\url{https://github.com/pytorch/fairseq/tree/main/examples/mbart}} dataset. For a sample $X^l$ in language $l$, mBART is trained to capture the monolingual knowledge during pretraining by optimizing follow objective:
\begin{eqnarray}
\label{eq:denoise}
\mathcal{L}_\mathrm{PT}=\sum_{l\in L}-\log P(X^l|N(X^l))
\end{eqnarray}
where $L$ represents the set of languages contained in CC-25, $\mathrm{N}(\cdot)$ is the noise function that mainly contains span masking and sentence permutation. { For span masking in mBART, $35\%$ of total words are selected, and each span is replaced with a mask token. The span length is determined by a Poisson distribution whose expected length is set to 3.5. } 

\paragraph{Standard Finetuning for Cross-Lingual Tasks}

Given a cross-lingual sentence pair $ S_i = \left (X_i, Y_i \right )$, standard finetuning directly feeds $X_i$ into the encoder and estimates parameters under the supervision of $Y_i$. 
As shown at the bottom of Figure~\ref{fig: main CS-Annealing}, we follow mBART to append source language ID and target language ID, e.g. ``[EN]'' and ``[ZH]'', to the end of $X_i$ and the beginning of $Y_i$. 
For the summarization task, besides language ID, we employ a symbol ``</s>'' to separate sentences. 
During finetuning downstream tasks, the model heavily depends on the cross-lingual information that multilingual Seq2Seq PLMs may lack, motivating us to introduce the cross-lingual information before standard finetuning.

\subsection{Two-Stage Finetuning with Code-Switching Restore Task}
\label{subsec:csrestore}

To sufficiently transfer monolingual knowledge in different languages to downstream cross-lingual tasks, we consider adopting a cross-lingual pretext task with \textit{easy-to-acquire cross-lingual knowledge}, i.e. ``Unsupervised Translation Vocabulary'' in the middle of Figure~\ref{fig: main CS-Annealing}, between multilingual Seq2Seq PLMs and standard finetuning.  

Following~\cite{artetxe2018acl}, we utilize unsupervised word embedding mapping to extract translation vocabulary with the \textit{downstream source-side or target-side monolingual corpus only}.
More specifically, given the downstream monolingual source and target corpus  denoted as $\mathcal {D}=\left \{ {S}_i=\left ( {x}_1,..,{x}_n; {y}_1,...,{y}_m \right ) \right \}$. We first train a word2vec~\cite{10.5555/2999792.2999959} model to get the in-domain word-level embedding for both source side and target side, i.e. $E_x$ and $E_y$. 
These two sets of embeddings are then embedded into a shared feature space via self-learning following. 
Thus, we can measure the semantic distance of two words in different languages by the dot product distance of the corresponding embedding pair. 
We randomly select one of the top-k nearest neighbors as the semantic similar words for switch operation. Formally, the selected neighbors for $x_i$ and $y_i$ are denoted as $x_i^\prime$ and $y_i^\prime$, respectively. 

\paragraph{The Code-Switching Restore Task} 
The code-switching restore task is designed for injecting the above unsupervised translation vocabulary to bridge the cross-lingual gap between pretraining and standard finetuning. 

The code-switching restore tasks will be performed on the downstream languages, we take the source-side code-switching as an example here, while operations are the same for the target language.
\begin{CJK}{UTF8}{gbsn}
As illustrated in Figure~\ref{fig: main CS-Annealing}, 
given a sentence pair $\mathcal {S}_i=\left (X_i, Y_i \right )=\left ({x}_1,..,{x}_n; {y}_1,..,{y}_m\right )$ selected from $\mathcal{D}$, where $X$ is Chinese sentence ``布什与沙龙举行了会谈$_{_{。}}$ '' and $Y$ is English sentence ``Bush held a talk with Sharon.'', 
we sample a text span ``举行了会谈'' from the Chinese sentence  $ {X}_i$ whose length is determined by a Poisson distribution. 
For each selected word $x_i$, e.g. ``举行'', we replace it with one of its unsupervised translation vocabulary $x_i^\prime$, i.e. ``held'', to derive the new code-switched source sentence.
Such sampling will be repeated until the total percentage of substituted words reaches $35\%$\footnote{We keep the substitution rate as the default mask ratio of mBART.}. Then the source sentence of our pretext task can be written as $ {X}_i^\prime=\left ( {x}_1,.., {x}_i^\prime,...,{x}_{i+j}^\prime,...,{x}_n \right )$, e.g. ``布什与沙龙 held a talk$_{_{。}}$ '' in our example. 

After that, we can get the pretext task sentence pair $\mathcal {S}_i = \left ( {X}_i^\prime; {X}_i \right )$, i.e. ``布什与沙龙 held a talk$_{_{。}}$ ''$\rightarrow$``布什与沙龙举行了会谈$_{_{。}}$ '' in our example. 
Then we feed this sentence pair into the multilingual Seq2Seq PLM, optimizing the cross-entropy loss. Considering the same operation for the target-side sentence $Y_j$, the total objective for the code-switching restore task can be written as:
\begin{eqnarray}
\begin{split}
\label{eq: CS restore} 
\mathcal{L}_\mathrm{pretext}=& \\
-\sum_{X_i \in \mathcal{D}} \log  P(X_{i}|X_{i}^\prime) &-\sum_{Y_j \in \mathcal{D}} \log  P( Y_{j}|Y_{j}^\prime)
\end{split}
\end{eqnarray}
where $X_i$, $Y_j$ mean source and target sentences in the downstream dataset, respectively, and $X_i^\prime$, $Y_j^\prime$ are correspondingly code-switched sentences.
\end{CJK}

\paragraph{Two-Stage Finetuning} 
After introducing downstream cross-lingual knowledge with our proposed code-switching restore task, the multilingual Seq2Seq PLMs will be easier to adapt downstream standard finetuning.
Thus, the recipe for using our approach follows a two-stage fashion:
In the \textbf{first} stage, we tune the multilingual Seq2Seq PLMs for certain steps using the above pretext task, where \textit{the model can gain a better in-domain initialization with the narrowed cross-lingual gap} for the downstream tasks.
In the \textbf{second} stage, we tune the model normally on the downstream task.

The effective performance of our method comes from the narrowed cross-lingual representation distance in \S\ref{subsec:CL_R_gap}, confirming our above \textit{claim}.

\begin{table*}[t]
\centering
\caption{\textbf{Performance of bilingual translation tasks} on WMT and IWSLT datasets. Results are in the format ``SacreBLEU / COMET-score''. (\textbf{Bold}: the best results. ``$^{\dagger/\ddagger}$'': significant statistical difference ($p<0.05/p<0.01$) from mBART.)}
\scalebox{1.2}{
\begin{tabular}{lcccccc}
\toprule
\bf Languages           & \multicolumn{2}{c}{\bf En-Vi}                 & \multicolumn{2}{c}{\bf En-De}                     & \multicolumn{2}{c}{\bf En-Tr} \\
\bf Data Source         & \multicolumn{2}{c}{\textbf{IWSLT15}}               & \multicolumn{2}{c}{\textbf{IWSLT14}}                  & \multicolumn{2}{c}{\textbf{WMT17}} \\
\bf Direction           & $\leftarrow$          & $\rightarrow$     & $\leftarrow$          & $\rightarrow$         & $\leftarrow$          & $\rightarrow$ \\ \midrule
\bf \textbf{Random}              & 23.6 / ~~~~-~~~~                  & 24.8 / ~~~~-~~~~              & 34.4 / ~~~~-~~~~                  & 29.2 / ~~~~-~~~~                  & 12.2 / ~~~~-~~~~                  & 9.5 / ~~~~-~~~~         \\
\bf \textbf{mBART}               & 28.2 / 0.150                  & 29.4 / 0.190              & 37.8 / 0.517                  & 32.6 / 0.413                  & 22.1 / 0.393                  & 20.0 / 0.823        \\
\sc \quad\bf +ours      & \bf~29.6$^\ddagger$/ 0.159   & \bf 29.7 / 0.213              & \bf 39.2$^\ddagger$/ 0.553   & 32.6 / \textbf{0.418}                  & \bf23.1$^\ddagger$/ 0.435   & \bf 20.9 $^\dagger$/ 0.872       \\ \midrule
\bf Languages           & \multicolumn{2}{c}{\bf En-Ko}                 & \multicolumn{2}{c}{\bf En-It}                     & \multicolumn{2}{c}{\bf En-Lt} \\
\bf Data Source         & \multicolumn{2}{c}{\textbf{IWSLT17}}               & \multicolumn{2}{c}{\textbf{IWSLT17}}                   & \multicolumn{2}{c}{\textbf{WMT19}} \\
\bf Direction           & $\leftarrow$          & $\rightarrow$     & $\leftarrow$          & $\rightarrow$         & $\leftarrow$          & $\rightarrow$ \\ \midrule
\bf \textbf{Random}              & 15.3 / ~~~~-~~~~                  & 16.3 / ~~~~-~~~~              & 31.7 / ~~~~-~~~~                  & 28.0 / ~~~~-~~~~                  & 18.1 / ~~~~-~~~~                  & 12.1 / ~~~~-~~~~        \\
\bf \textbf{mBART}               & 17.2 / 0.141              & 33.4 / 0.273              & 37.3 / 0.434                 & 34.8 / 0.534                  & 29.1 / 0.488                  & 12.8 / 0.302        \\
\sc \quad\bf +ours      & \bf 18.5$^\ddagger$/ 0.202   & \bf 33.5 / 0.301              & \bf 38.8$^\ddagger$/ 0.452   & \bf 35.7$^\dagger$/ 0.572   & \bf 30.8$^\ddagger$/ 0.507   & \bf 13.5$^\dagger$/ 0.316       \\
\bottomrule
\end{tabular}}
\label{tab:main}
\end{table*}

\section{Experiments}
\label{sec: exp}
We conduct cross-lingual \textbf{NLG} evaluations on four typical tasks: \romannumeral1) {bilingual translation}; \romannumeral2) {zero-shot translation}; \romannumeral3) {cross-lingual summarization}, and \textbf{NLU} evaluations on representative { iv) cross-lingual Natural Language Inference (cross-lingual NLI)} tasks.

\subsection{Experimental Setup}
{ \paragraph{Baselines} We select three models as our baseline models for comparison. 
For cross-lingual generation tasks, we select Random and mBART to verify performance improvement of our approach. And, mBERT and mBART are used for cross-lingual NLI experiments. } 
\begin{itemize}
    \item \textbf{Random}: the Transformer model~\cite{transformer} whose parameter is randomly initialized, and trained from scratch. 
    { \item \textbf{mBERT}: multilungual BERT~\cite{devlin-etal-2019-bert} indicates the bidirectional transformer pretrained on the Wikipedias of 104 languages using Masked Language Model (MLM) task, finetuning on cross-lingual NLI tasks.  }
    {\item \textbf{mBART}: directly finetuning the mBART~\cite{tacl_mBART} model on downstream tasks with identical update steps to our proposed two-stage finetuning strategy. To obtain the best performance, we apply grid search to get hyper-parameters, e.g. \textit{lr, batch size, training steps}, on the validation set.}
    
\end{itemize}

\paragraph{Model Training} 
{ We select mBART~\cite{tacl_mBART}, the multilingual sequence-to-sequence pretrained language model, pretrained on CC-25 dataset as our backbone model. }
For each task, source and target languages are jointly tokenized into sub-word units with the 250k SentencePiece~\cite{kudo-richardson-2018-sentencepiece} vocabulary of mBART~\cite{tacl_mBART}. 
All experiments are conducted on the open-source toolkit {\tt fairseq}~\cite{ott-etal-2019-fairseq}. We set hyperparameters as follows, including 0.3 as the dropout ratio, 0.2 as the label smoothing ratio, 2500 as warm-up steps, and 3e-5 as the learning rate. The maximum update steps for different tasks are presented in following sections. Unless otherwise stated, we use beam search with size 5 in decoding.

For unsupervised translation vocabulary acquisition, we closely follow~\cite{artetxe2018acl} as described in \ref{subsec:csrestore}. Recall that the word embeddings for code-switching are only trained using datasets of downstream tasks.

\subsection{Bilingual Translation} 
\label{sec:Bilingual Translation} 

The bilingual translation task aims to translate a source language sentence into the target language. 
In this task, we conducted 12 experiments on publicly available parallel corpora to evaluate the superiority of our approach. Two of these corpora are obtained from WMT (Tr, Lt $\leftrightarrow$ En), while others are gathered from IWSLT (Vi, De, Ko, It $\leftrightarrow$ En). For data pre-processing, we follow~\cite{tang-etal-2021-multilingual} to remove duplicated sentence pairs.
{We set the training steps to 5k for the pretext task, and to 25k for the translation task, respectively.}
For evaluation, we use sacreBLEU~\cite{post-2018-call} to measure the translation quality. 
{ We select Mecab-Ko with Hannanum\footnote{\url{http://konlpy.org/en/v0.3.0/install/}} to perform segmentation when evaluate the performance for ko generation. As for other languages, we used the default tokenizer of official SacreBLEU. }
We also use COMET\footnote{\tt wmt-large-da-estimator-1719}~\cite{rei-etal-2020-comet} to evaluate semantic faithfulness, which is more correlated to human judgments. It is worth to notify that, for supervised models, we directly report the SOTA random initialized results from mBART and previous works.

Table~\ref{tab:main} lists the results. 
We can see that our approach achieves better translation than both supervised and mBART baselines in all languages.
Noticeably, our method achieves improvements against mBART using standard finetuning on 9 out of 12 directions with sufficient significance, proving the effectiveness of introducing a code-switching restore task to bridge the cross-lingual gap for multilingual Seq2Seq PLM. 
We also see XX$\rightarrow$En translation tasks all achieve sufficient performance improvement, which indicates our approach is relatively more beneficial for English as target language.
{ The results of mBART consistently outperforms the randomly initialized transformer by a large margin, which  confirms the benefit of pretraining and is agree with previous works.

Table~\ref{tab:main} also shows that our approach can achieve consistent COMET-score improvements compared to standard finetuning across 12 experiments. We got an extra promotion of up to 0.061 (Ko$\rightarrow$En). A higher COMET score means that the sentences translated by our approach are more in line with human translation.
}

\subsection{Zero-Shot Translation} 
\begin{table*}[t]
\centering
\caption{\textbf{Performance of zero-shot translation tasks} on XX$\rightarrow$En directions. Results are in the format ``SacreBLEU / COMET-score''.
(\textbf{Bold}: the best results. ``$^{\dagger/\ddagger}$'': significant statistical difference ($p<0.05/p<0.01$) from mBART.)}
\scalebox{1.2}{
\begin{tabular}{ccc|cccccc}
\toprule
&                     &       & \multicolumn{6}{c}{\bf Finetuned Languages}      \\
&                     &       & \textbf{Vi}   & \textbf{De}   & \textbf{Tr}   & \textbf{Ko}   & \textbf{It}   & \textbf{Lt}   \\
\midrule
\multirow{12}{*}{\STAB{\rotatebox[origin=c]{90}{\bf Testing Languages}}} 
    & \multirow{2}{*}{\textbf{Vi}} & \textbf{mBART}                 & 28.2 / 0.150                        & 15.3 /-0.119                      & 15.5 /-0.174                      & 18.5 /-0.095                      & 19.0 /-0.101                      & 14.0 /-0.291 \\
    &                     & \sc \quad\bf{+ours}   & \bf29.6 / \bf 0.159                     & 15.2 /\bf-0.116                      & \bf15.8 /\bf-0.147                   & \bf18.6 / \bf0.078                   & 19.0 /\bf-0.065                      & 13.6 /-0.322 \\\cline{2-9}
    & \multirow{2}{*}{\textbf{De}} & \textbf{mBART}                 & 22.0 / 0.091                        & 37.8 / 0.517                      & 19.9 / 0.100                      & 23.2 / 0.133                      & 24.2 / 0.097                      & 20.4 /-0.004 \\
    &                     & \sc \quad\bf{+ours}   & 22.0 / \textbf{0.107}               & \bf39.2$^\ddagger$/ \bf0.553         & \bf20.8$^\dagger$/ \bf0.139          & \bf23.6 / \bf0.169                   & 24.2 / \textbf{0.121}                      & \textbf{20.5} /-0.044 \\\cline{2-9}
    & \multirow{2}{*}{\textbf{Tr}} & \textbf{mBART}                 & ~8.8~ /-0.212                       & ~9.0~ /-0.111                     & 22.1 / 0.393                      & 15.0 / 0.089                      & ~9.1~ /-0.189                     & 13.4 / 0.023 \\
    &                     & \sc \quad\bf{+ours}   & \bf~9.0~ /-0.134                    & ~8.7~ /\textbf{-0.077}                     & \bf23.1$^\ddagger$/ \bf0.435         & \bf15.4 / \bf0.153                   & \bf~9.7$^\dagger$~/\bf-0.107         & \textbf{13.9}$^\dagger$/-0.002 \\\cline{2-9}
    & \multirow{2}{*}{\textbf{Ko}} & \textbf{mBART}                 & ~7.7~ /-0.326                       & ~8.0~ /-0.270                     & ~8.8~ /-0.154                     & 17.2 / 0.141                      & ~6.8~ /-0.375                     & ~7.1~ /-0.234 \\
    &                     & \sc \quad\bf{+ours}   & \bf~7.9~ /\bf-0.242                    & \bf~8.3~ /\bf-0.191                  & \bf~9.7$^\dagger$~/\bf-0.104         & \bf18.5$^\ddagger$/ \bf0.202         & \bf~7.9$^\ddagger$~/\bf-0.219        & \textbf{~8.3}$^\ddagger$~/\bf-0.189\\\cline{2-9}
    & \multirow{2}{*}{\textbf{It}} & \textbf{mBART}                 & 28.4 / 0.087                        & 24.9 / 0.017                      & 23.8 / 0.096                      & 27.4 / 0.161                      & 37.3 / 0.434                      & 23.7 /-0.122 \\
    &                     & \sc \quad\bf{+ours}   & \bf29.2$^\ddagger$/ \bf0.121          & \bf25.6$^\ddagger$/ \bf0.053          & \bf24.7$^\ddagger$/ \bf0.123         & \bf27.8 / \bf0.183                   & \bf38.8$^\ddagger$/ \bf0.452         & 23.5 /-0.146 \\\cline{2-9}
    & \multirow{2}{*}{\textbf{Lt}} & \textbf{mBART}                 & 12.9 /-0.093                        & 11.5 /-0.784                      & 18.2 / 0.133                      & 14.9 /-0.012                      & 12.7 /-0.113                      & 29.1 / 0.488 \\
    &                     & \sc \quad\bf{+ours}   & \bf13.5$^\dagger$/ \bf0.003            & 11.4 / \textbf{0.017}                      & 17.8 / \textbf{0.160}                      & \bf15.2 / \bf0.069                   & \bf13.5$^\dagger$/ \bf0.010          & \bf30.8$^\ddagger$/ \bf0.507 \\
\bottomrule
\end{tabular}}
\label{tab:UNMT with transfer}
\end{table*} 
The zero-shot translation tasks aim to transfer the ability of pretrained YY$\rightarrow$En model to the zero-resource language pairs XX$\rightarrow$En.

Generally, the closer distance between language XX and YY, the better transfer performance will be realized by the model.
In our settings, we use all the above 6 bilingual translation corpora, and directly utilize the models finetuned on YY$\rightarrow$En from bilingual translation experiments in \S\ref{sec:Bilingual Translation} to test on XX to En. 
We consider the same metrics in the bilingual translation tasks for evaluation, i.e. SacreBLEU and COMET. 

As shown in Table~\ref{tab:UNMT with transfer}, we present the zero-shot results of mBART models under the standard finetuning pattern and under our proposed approach. Obviously, our method shows improvements on 21 (in terms of BLEU scores) and 26 (in terms of COMET scores) out of 30 zero-shot directions, demonstrating the universality and robustness of our method.

\subsection{Cross-Lingual Summarization}

\begin{table}[t]
\centering
\caption{\textbf{Performance of cross-lingual summarization tasks} on NCLS dataset. 
The best results are \textbf{bold}. }
\scalebox{1.2}{
\begin{tabular}{lccc}
\toprule
\textbf{} & \textbf{Rouge-1}  & \textbf{Rouge-2}  & \textbf{Rouge-L}  \\
\hline
\textbf{} & \multicolumn{3}{c}{\bf En$\rightarrow$Zh}  \\ \hdashline
\bf \textbf{Random} & ~~6.5 & ~1.1~ & ~~6.4 \\
\bf \textbf{mBART} & 22.1 & ~8.7~ & 21.7 \\
\sc \quad\bf+ours & \bf22.4 & \bf~9.1~ & \bf22.1 \\ \midrule
\textbf{} & \multicolumn{3}{c}{\bf Zh$\rightarrow$En}  \\ \hdashline
\bf \textbf{Random} & 20.8 & ~~6.0 &  16.5 \\ 
\bf \textbf{mBART} & 27.1 & 11.4 &  22.5 \\   
\sc \quad\bf+ours & \bf28.2 & \bf12.1 & \bf 23.4 \\ 
\bottomrule
\end{tabular}}

\label{tab:cross-lingual experimental results}
\end{table} 

The cross-lingual summarization task aims to automatically generate a target language summary that retains the most important content of the source language document.
We conduct the cross-lingual summarization experiments on NCLS dataset~\cite{zhu-etal-2019-ncls}, which contains 370K En$\rightarrow$Zh document-summary pairs and 1.69M Zh$\rightarrow$En document-summary pairs.
 
For too-long documents, we follow~\cite{zhu-etal-2019-ncls} to truncate the input sentence to 220 words, and the output to 130 words (160 words for Chinese output).
{We empirically employ different batch sizes for En$\rightarrow$Zh and Zh$\rightarrow$En directions (49K and 98K) according to their document lengths, i.e. En$\rightarrow$Zh is significantly larger than that of Zh$\rightarrow$En.
We set the training steps to 5k and 50k for the pretext task and the rest cross-lingual summarization task, respectively.
After finetuning, the model could generate the English/ Chinese summarization according to the Chinese/ English document.} 
We follow~\cite{lin-2004-rouge}\footnote{\url{https://github.com/pltrdy/files2rouge}} to report the F1-score of ROUGE-1, ROUGE-2, and ROUGE-L. The summaries are generated by beam search with size$=$6 in decoding. { After generation, we use Standford NLP tokenizer\footnote{\url{https://nlp.stanford.edu/software/tokenizer.shtml}} to segment words for English, while Chinese texts are split by characters.}

Table~\ref{tab:cross-lingual experimental results} lists the performance of generated summarization. Clearly, our model outperforms other baselines consistently. Compared with the model after the standard finetuning, our model achieves +0.4 in ROUGE-L for En$\rightarrow$Zh, and +0.9 in ROUGE-L for Zh$\rightarrow$En, indicating the superiority of our method.

\subsection{{Cross-Lingual NLI}}
\begin{table}[t]
\centering
\caption{\textbf{Performance of cross-lingual NLI tasks} on datasets from XNLI. Average scores on all tasks are \underline{underlined}. The best results are \textbf{bold}.}
\scalebox{1.2}{
\begin{tabular}{lcccc}
\toprule
\multicolumn{1}{l}{} & \textbf{De} & \textbf{Es} & \textbf{Fr} & \textbf{Ru} \\ 
\hline
\textbf{mBERT} & 71.1 & 74.3 & 73.8 & 69.0 \\
\textbf{mBART} & 76.4 & 79.0 & 78.4 & 74.0 \\
\quad\bf +ours & \textbf{77.3} & \textbf{79.4} & \textbf{79.0} & \textbf{75.2} \\
\toprule
\multicolumn{1}{l}{} & \textbf{Tr} & \textbf{Vi} & \textbf{Zh} & \textit{\textbf{Avg.}} \\
\hline
\textbf{mBERT} & 61.6 & 69.5 & 69.3 & \underline{69.8} \\
\textbf{mBART} & 68.2 & 74.7 & 69.6 & \underline{74.3} \\
\quad\bf +ours & \textbf{71.9} & \textbf{76.1} & \textbf{71.2} & \underline{\textbf{75.7}} \\ 
\bottomrule
\end{tabular}}
\label{tab: XNLI results}
\end{table}

In the cross-lingual natural language inference (cross-lingual NLI) task, we finetune the model on English training data and perform inference on other languages. 
We use datasets from XNLI~\cite{conneau-etal-2018-xnli} benchmark and report the accuracy.
In this task, we utilize the publicly accessible bilingual embeddings MUSE\footnote{\url{https://github.com/facebookresearch/MUSE}}~\cite{lample2018word} to imply { code-switching}.
For model finetuning, we set the batch size to 128, the learning rate to 5e-6 for 10 epochs iteration, performing early stopping on the validation set.

Table~\ref{tab: XNLI results} shows that our proposed finetuning approach outperforms the other two strong baselines, i.e. mBERT\footnote{We report the results of mBERT from~\cite{liang2020xglue}.}~\cite{Devlin2019BERTPO} and mBART with standard finetuning. Our model achieves +1.4 average score improvement over mBART and +5.9 over mBERT. Our method achieves the improvement by a large margin on language like ``Tr'' whose pretraining corpus is limited in mBART, showing the transferability and effectiveness of our proposed method on NLU tasks, especially for low-resource languages.

\section{Discussion}
\label{sec: discussion}
To better understand how our approach alleviates the cross-lingual gaps between multilingual Seq2Seq PLMs and downstream tasks, we make in-depth analyses to investigate three problems: 
\textbf{Q1:} How does our method narrow cross-lingual representation gaps? (\S\ref{subsec:CL_R_gap}) 
\textbf{Q2:} Whether our approach improves the low-frequency word translation? (\S\ref{subsec:LF_W}) 
and \textbf{Q3:} Does our strategy acquire large computational costs? (\S\ref{subsec:computational_cost}) 

\subsection{Narrowing Cross-Lingual Representation Distance}
\label{subsec:CL_R_gap}
\begin{table}[t]
\setlength{\belowcaptionskip} {-0.3cm}
\centering
\caption{\textbf{The distances of cross-lingual sentence representations} on language directions En-XX. The closet distances are \textbf{bold}.}
\scalebox{1.2}{
\begin{tabular}{lcccc}
\toprule
\multicolumn{1}{r}{\textbf{Language}} & \textbf{De} & \textbf{Ko} & \textbf{Vi} & \textbf{Lt} \\ \midrule
\bf Pretrained & 28.33 & 90.64 & 64.84 & 36.68 \\
\bf Finetuned & ~~8.25 & ~~7.22 & ~~8.38 & 11.37 \\
\sc \quad\bf+ours & \bf ~~7.90 & \bf ~~4.80 & \bf ~~6.70 & \bf ~~5.17 \\
\bottomrule
\end{tabular}}
\label{tab:narrowing cross-lingual gap}
\end{table}

Many previous works have shown that narrowing the cross-lingual gaps can bring better modeling performance~\cite{Zhou2019HandlingSD,ding-etal-2020-self}.
To quantify the narrowed cross-lingual representation distance, we take the average sentence embedding to represent each language and evaluate the language distance on CC-100~\cite{conneau-etal-2020-unsupervised}. In practice, we select a subset $S$ containing 20k sentences for approximation.

Given a sentence $\mathcal {S}_i=\left ( {x}_1,..,{x}_n \right )$, we feed it into the model to obtain the sentence embedding ${e}_i$, where ${e}_i$ is the output on the last token from the decoder. Applying the Euclidean distance, the distance between the source and target language is therefore defined as:
\begin{eqnarray}
\label{eq:language distance} 
{\mathcal{D}_\mathrm{language}=(\sum_{i=1}^{20k}\left|e_i^{source}-e_i^{target}\right|^2)^{1/2}}
\end{eqnarray}

{Table~\ref{tab:narrowing cross-lingual gap} shows the language distances of four language pairs. We notice that the cross-lingual representation distances are reduced in all scenarios after finetuning. Moreover, our method can further reduce the distance up to -6.20 reduction on En-Lt. 
This demonstrates that our strategy could improve the finetuning performance by significantly reducing the cross-lingual representation distance.
}

\subsection{Improving Low-Frequency Word Translation}
\label{subsec:LF_W} 

\begin{table}[t]
\centering
\caption{\textbf{Performance of word translation accuracy with different frequencies} on IWSLT14 De$\rightarrow$En test set. ``Low'', ``Mid'' and ``high'' denote low-, mid-, and high-frequency words, respectively. 
}
\scalebox{1.2}{
\begin{tabular}{lcccc}
\toprule
\multicolumn{1}{l}{\textbf{Model}}   & \textbf{All} & \textbf{High} & \textbf{Mid} & \textbf{Low} \\ \hline
\bf Finetuned          & 63.2       & 68.7   &58.4 & 48.6     \\
\sc \quad\bf+ours       & \bf63.5        & 68.7   &  \bf 58.8 & \bf 50.1   \\
\bottomrule
\end{tabular}}
\label{tab: LF_W}
\end{table} 

\begin{table*}[t]
\centering
\caption{\textbf{Case Study for De$\rightarrow$En translation.} We present translation examples for low-frequency and med-frequency words, which are marked with \underline{underline}. ``Input'' and ``Human'' indicate the source German sentences and the ground-truth target English sentences, respectively. }
\begin{tabular}{p{1.2cm}p{6.3cm}p{1.2cm}p{6.3cm}}
\toprule
 & \textbf{Sentence}          &       & \textbf{Sentence} \\ \hline
\textbf{Input}                & nichts wandert in meinen ärmel oder kommt heraus, keine trickserei. und sie können alles untersuchen.  & \textbf{Input} & passiert ist das das so gennante maschinenzeitalter. \\ \hdashline
\textbf{Human}                & nothing goes up or down my \underline{sleeve}, no \underline{trickery}. and you can \underline{examine} everything. & \textbf{Human} & what happened is the the \underline{so-called} machine age. \\
\textbf{mBART}                & nothing goes into my throat or comes out, no tricks, and you can study everything.   & \textbf{mBART} & this is the so genetic age of machines. \\
\textbf{Ours}                 & nothing goes into my sleeve or comes out of my \underline{sleeve}, no \underline{trickery}. and you can \underline{examine} anything. & \textbf{Ours}  & this is the \underline{so-called} machine age.\\ \midrule
 & \textbf{Sentence}                       &     & \textbf{Sentence} \\ \hline
\textbf{Input}                & es gibt nur eine und das sind die vereinigten staaten -- zum glück, zum glück. & \textbf{Input} & dies ist, denke ich, so tief in unserer wasserversorgung verankert, dass es keinem in dem sinn kommen würde es zu hinterfragen. \\ \hdashline
\textbf{Human}                & there is only one, and that's the united states -- \underline{fortunately}, \underline{fortunately}.  & \textbf{Human} &    this, i think, is so deeply \underline{embedded} in the water supply that it wouldn't \underline{occur} to anyone to question it.   \\
\textbf{mBART}                & there's only one, and that's the united states -- fortunate, fortunate.                         & \textbf{mBART} &   this is so deeply rooted in our water supply, i think, that nobody would think of it as questioning it.   \\
\textbf{Ours}                 & there's only one, and that's the united states -- \underline{fortunately}, \underline{fortunately}.   & \textbf{Ours}  &  this is so deeply \underline{embedded} in our water supply, i think, that it would never \underline{occur} to anybody to question it.  \\ 
\bottomrule
\end{tabular}
\label{tab: case-study}
\end{table*}

Each selected word in our code-switching restore stage will be replaced with multiple candidates in another language, thus our method is expected to obtain better generalization, especially at the word level. 
To verify our hypothesis, we measure the translation performance of low-frequency words, {  which evaluated with F-measure by using compare-mt\footnote{\url{https://github.com/neulab/compare-mt}}. We calculate word frequency on the train set.}
The word-level translation performance in terms of F-scores on the IWSLT14 De$\rightarrow$En test set is shown in Table~\ref{tab: LF_W}. 
Words that appear less than 100 times are regarded as low-frequency words while more than 1000 times are regarded as high-frequency words, and the rest belong to mid-frequency ones. 

We see that our approach boosts the translation accuracy, especially for low-frequency words where a more remarkable improvement can be observed, i.e. +1.5, nicely validating our hypothesis. { We further present the case study for IESLT14 De$\rightarrow$En translation results in Table~\ref{tab: case-study}. Compared with standard fine-tuning, our model gives low-frequency word translations that are more agree with human translations.}

\subsection{Trivial Computational Costs}
\label{subsec:computational_cost}
% \begin{figure}[t]
% \begin{tikzpicture}
% % \small
%  \begin{axis}[
%   width=1\columnwidth,
%   height=0.9\columnwidth,
%   legend cell align=left,
%   legend style={at={(1, 1)},anchor=north east,font=\normalsize},
%   mark options={mark size=4},
%   font=\normalsize,
%   xmin=0, xmax=25,
%   ymin=5, ymax=10,
%   xtick={5,10,15,20,25},
%   ymajorgrids=true,
%   xmajorgrids=true,
%   xlabel style={yshift=0.5ex,},
%   xlabel=Finetuning Steps (K) ,
%   ylabel=Training Loss , 
%   ylabel style={yshift=-0.5ex,}]
%     \addplot[mark=triangle,g-blue,line width=1pt] plot coordinates {
%       (0.61, 14.940)
%       (2.457, 7.114)
%       (4.921, 6.269)
%       (9.849, 5.891)
%       (14.777, 5.761)
%       (19.705, 5.588)
%       (25.000, 5.523)
%     };
%     \addlegendentry{ Standard Finetuning}
%     \addplot[mark=x,g-red,line width=1pt] plot coordinates {
%       (0.61, 7.698)
%       (2.457, 6.286)
%       (4.921, 5.883)
%       (9.849, 5.598)
%       (14.777, 5.458)
%       (19.705, 5.378)
%       (25.000, 5.339)
%     };
%     \addlegendentry{Ours Approach}
%  \end{axis}
% \end{tikzpicture}
% \caption{\textbf{Comparison of training dynamics} during finetuning on IWSLT14 De$\rightarrow$En dataset. For ``Ours Approach'', we present the loss curve of the second stage. 
% % The x-axis is the value of $k$.
% }
% \label{fig:step_abalation}
% \end{figure}

% TASLP version
\begin{figure}[t]
\centering
\begin{tikzpicture}
\small 
 \begin{axis}[
   width=0.86\columnwidth,
   height=0.8\columnwidth,
   legend cell align=left,
   legend style={at={(1, 1)},anchor=north east,font=\normalsize},
   mark options={mark size=4},
   font=\normalsize,
   xmin=0, xmax=25,
   ymin=5, ymax=10,
   xtick={5,10,15,20,25},
   ymajorgrids=true,
   xmajorgrids=true,
   xlabel style={yshift=0.5ex,},
   xlabel=Finetuning Steps (K) ,
   ylabel=Training Loss , 
   ylabel style={yshift=-0.5ex,}]
    \addplot[mark=triangle,g-blue,line width=1pt] plot coordinates {
      (0.61, 14.940)
      (2.457, 7.114)
      (4.921, 6.269)
      (9.849, 5.891)
      (14.777, 5.761)
      (19.705, 5.588)
      (25.000, 5.523)
    };
    \addlegendentry{ Standard Finetuning}
    \addplot[mark=x,g-red,line width=1pt] plot coordinates {
      (0.61, 7.698)
      (2.457, 6.286)
      (4.921, 5.883)
      (9.849, 5.598)
      (14.777, 5.458)
      (19.705, 5.378)
      (25.000, 5.339)
    };
    \addlegendentry{Ours Approach}
 \end{axis}
\end{tikzpicture}
\caption{\textbf{Comparison of training dynamics} during finetuning on IWSLT14 De$\rightarrow$En dataset. For ``Ours Approach'', we present the loss curve of the second stage. 
% The x-axis is the value of $k$.
}
\label{fig:step_abalation}
\end{figure}
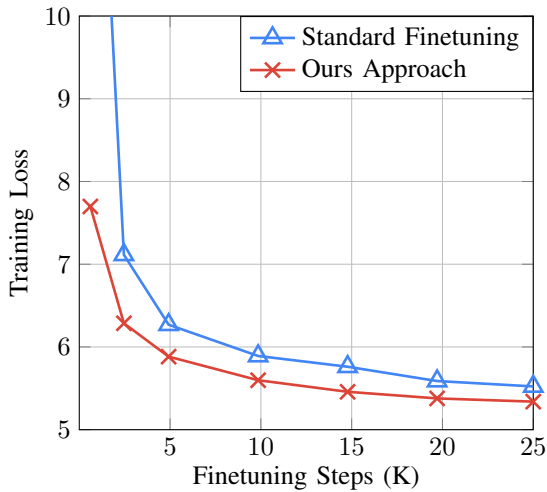
As shown in Figure~\ref{fig:step_abalation}, we also conclude that the code-switching restore task benefits the finetuning converges speed. 
At the second finetuning stage of our method, the model after the pretext task only requires about half of the update steps (12k vs. 25k) to reach the same loss level, compared to standard finetuning. 
Since the tuning step number in the first stage is set to 5k (only 16.7\% of total steps), we argue that the computational costs are not an obstacle to the extensibility of our approach. 
Additionally, our model can achieve even lower loss with further training applied, indicating that the code-switching restore task can adapt the model in advance to cross-lingual downstream tasks more effectively.

\section{Conclusion} 
\label{sec:conclusion}
In this work, {we propose a feasible code-switching based finetuning approach} to mitigate the cross-lingual gap in the multilingual Seq2Seq pretrain-finetune paradigm. In our method, a model needs to reconstruct a sequence in the source language with part of words replaced by their neighbors from the unsupervised translation vocabulary in the target language, and vice versa.
Experimental results illustrate the effectiveness and universality of our finetuning approach on both cross-lingual generation and language understanding tasks. 
Moreover, extensive analyses also prove that our pretext task can significantly narrow the cross-lingual representation distance and boost the low-frequency word translation with trivial computational cost.

In future work, we will explore the effectiveness of our strategy on more multilingual Seq2Seq PLMs, e.g. mT5~\cite{xue-etal-2021-mt5}, and supervised multilingual Seq2Seq pretrained translation models~\cite{lin2020pre,facebook2021,jdexplore2022}. Meanwhile, it will be interesting to design more efficient pretext tasks to further reduce the cross-lingual gaps in multilingual Seq2Seq PLMs.
 
\bibliography{main}
\bibliographystyle{IEEEtran}

\end{document}